\pdfoutput=1
\documentclass[conference,letterpaper,10pt,final.hidelinks]{ieeeconf}
\usepackage[utf8]{inputenc}
\usepackage[T1]{fontenc}
\usepackage{graphicx}
\usepackage{longtable}
\usepackage{wrapfig}
\usepackage{rotating}
\usepackage[normalem]{ulem}
\usepackage{amsmath}
\usepackage{amssymb}
\usepackage{capt-of}
\IEEEoverridecommandlockouts
\usepackage{algorithm,algpseudocodex,soul,amsmath}
\usepackage[font=small]{caption}
\usepackage{arydshln}
\usepackage{booktabs,multirow}
\usepackage{subfig}
\usepackage[utf8]{inputenc}
\usepackage[T1]{fontenc}
\usepackage{ulem}
\usepackage[colorinlistoftodos, textwidth=3cm]{todonotes}
\usepackage[hidelinks]{hyperref}
\date{}
\title{Informative path planning for scalar dynamic reconstruction using coregionalized Gaussian processes and a spatiotemporal kernel}
\hypersetup{
 pdfauthor={Lorenzo Booth and Stefano Carpin},
 pdftitle={Informative path planning for scalar dynamic reconstruction using coregionalized Gaussian processes and a spatiotemporal kernel},
 pdfkeywords={},
 pdfsubject={},
 pdfcreator={Emacs 30.0.50 (Org mode 9.6.6)},
 pdflang={English}}
\begin{document}

% Fix for inline todos (todonotes package)
\makeatletter
\let\orgtitle\@title
\if@todonotes@disabled
\newcommand{\hlfix}[2]{#1}
\else
\newcommand{\hlfix}[2]{\texthl{#1}\todo{#2}}
\fi

% red strkeouts
\newcommand\redsout{\bgroup\markoverwith{\textcolor{red}{\rule[0.5ex]{2pt}{0.4pt}}}\ULon}

% Enable lines in matrix environment (amsmath)
\renewcommand*\env@matrix[1][*\c@MaxMatrixCols c]{%
  \hskip -\arraycolsep
  \let\@ifnextchar\new@ifnextchar
  \array{#1}}

% Override ieeeconf smallcaps in captions
\patchcmd{\@makecaption}
  {\scshape}
  {}
  {}
  {}
\makeatother

\title{\bf \orgtitle}
\author{Lorenzo Booth \qquad Stefano~Carpin% <-this % stops a space
\thanks{\noindent  L.~Booth and S.~Carpin are with the Department of Computer Science and Engineering, University of California, Merced, CA, USA. L. Booth is supported by the Labor \& Automation in California Agriculture (LACA) project, which is part of the University of California's Office of the President Multicampus Research Programs \& Initiatives. This work was partially funded by NSF CBET-1604906. S. Carpin is partially supported by the IoT4Ag Engineering Research Center funded by the National Science Foundation (NSF) under NSF Cooperative Agreement Number EEC-1941529.}}

\maketitle

\begin{abstract}
The proliferation of unmanned vehicles offers many opportunities for solving
environmental sampling tasks with applications in resource monitoring and
precision agriculture. Informative path planning (IPP) includes a family of
methods which offer improvements over traditional surveying techniques for
suggesting locations for observation collection. In this work, we present a
novel solution to the IPP problem by using a coregionalized Gaussian processes
to estimate a dynamic scalar field that varies in space and time. Our method
improves previous approaches by using a composite kernel accounting for
spatiotemporal correlations and at the same time, can be readily incorporated in
existing IPP algorithms. Through extensive simulations, we show that our novel
modeling approach leads to more accurate estimations when compared with formerly
proposed methods that do not account for the temporal dimension.
\end{abstract}

\section{Introduction}
\label{sec:org8b7238c}

Consider the task of modeling a soil property in an agricultural field with a
point sensor. Whether the sensor is wielded by a human or an autonomous robot,
the agent is tasked with deciding where to capture observations of the
environment in order to inform the spatial interpolation. If the environmental
properties are dynamic and can change over the course of the survey, the
operator is also tasked with the option of updating an old measurement at a
previously-visited site, or measuring an unvisited location. When sampling under
practical constraints such as time and fuel, the operator must strategically
choose sampling locations that allow for useful predictive ability in space and
time, in order to arrive at a cohesive estimation of the system's state at the
end of the survey.

Thus, this task of \emph{informative path planning} (IPP) shares many elements with
the task of \emph{optimal sensor placement} and can be formalized as a \emph{constrained
optimization}, where the agent must evaluate the best location to travel, to
satisfy an objective function based in reconstructing a spatial process
\cite{krauseetal2008_nearoptimalsensorplacements,binneysukhatme2012_branchboundinformative}.
Recently, there have been many improvements in approaches to the IPP task for
various objectives including: map reconstruction with distributed agents, source
position estimation for sound and contaminant plumes, and search and rescue
\cite{popovicetal2020_informativepathplanninga}.

% Figure caption space is 10pt by default, here we narrow it to 5
%\vspace{-10pt}
\begin{figure}[h]
    \centering
    \includegraphics[width=1.0\columnwidth]{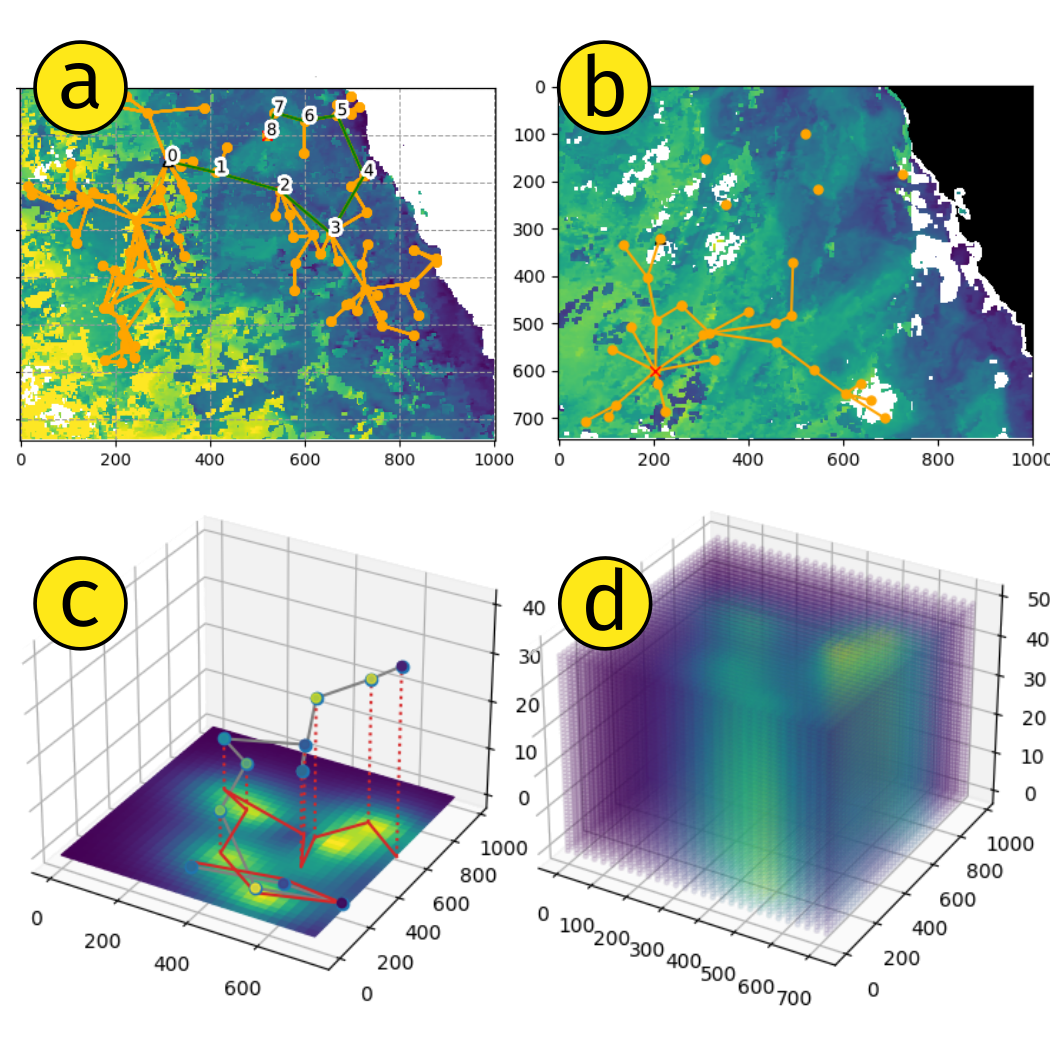}
     \vspace{-10pt}
    \caption{An overview of our evaluation methodology. (a) shows the ground
    truth, and the vehicle in the replanning stage, with observation history
    enumerated. (b) shows the environment during the planning stage with the
    locations of previous observations. (c) Samples can be visualized along a
    path in a temporal dimension and (d) displays the final map estimate at
    all inducing points in the Gaussian process.
    }
    \label{fig1}
\vspace{-20pt}
\end{figure}

Steady efforts have been directed toward sensing strategies for monitoring
spatiotemporal processes
\cite{caleyhollinger2015_datadrivencomparisonspatiotemporal}. The emergence of
small, inexpensive mobile platforms points to a future where mobile sensors may
be rapidly dispatched to model a dynamic phenomenon. However, to the best of our
knowledge there have been limited investigations of informative planners that
consider the \emph{temporal dimension} of information content, especially in an
online planning approach. This is necessary to produce faithful representations
of dynamic environments, as observations made early in the course of a survey
may no longer represent the state of the system at the location at the end of
the survey. Additionally, it may be desirable to infer the state of the system
at arbitrary points in time, or into the future.

To address this issue, we propose a novel sampling-based IPP framework that
considers the information content of sensing locations in space and time. An
overview of the framework is shown in Figure \ref{fig1} and in the accompanying
video. Inspired by the asymptotic optimality of IPP methods based on random
trees \cite{hollingersukhatme2014_samplingbasedroboticinformation}
\cite{jadidietal2019_samplingbasedincrementalinformation} and advancements in
large-scale, multiple-output Gaussian process modeling
\cite{hamelijncketal2021_spatiotemporalvariationalgaussiana}, our method
combines an information-theoretic sampling-based planner with a spatiotemporal
covariance function implemented as a separable kernel to access the information
gain from the locations of candidate sensing locations both in space \emph{and} time.
This also allows for both inference of the state and inference of model
uncertainty for unexplored parts of the system and establishes a criterion for
revisiting already-observed locations that no longer meaningfully reduce
uncertainty of the system's current state.

The contributions of this work are:
\begin{itemize}
\item A framework for reasoning about the information content of observations in
arbitrary dimensions reconciled to a metric appropriate for path planning
\item The integration of this spatiotemporal information function in a novel
time-aware informative planner for terrestrial monitoring
\item Validation of the approach in the context of spatial and temporal priors
with simulated and real-world dynamic scenarios inspired by common environmental
dispersion processes
\item Exploration of interactions between the parameters governing the planner and
the model
\end{itemize}

Our work opens up several avenues for consideration: the continuous update of
spatial and temporal priors through adaptive planning, extensions into
multi-robot systems, combined sensing modalities for prediction in multiple
dimensions (in a manner similar to Co-Kriging in the geostatistical literature),
and extensions into different classes of multi-output Gaussian processes. Our
framework will be open-sourced, for use in future investigations.

This paper is organized as follows: Selected related work is presented in
\autoref{related}. The problem formulation is introduced in \autoref{statement} and
our methods are discussed in \autoref{methods}. In \autoref{evaluation} we
experimentally evaluate our proposal and conclude in \autoref{conclusion}.

\section{Related Work}
\label{related}
This paper draws from a rich body of literature, surrounding the task of
collecting observations by an autonomous agent for modeling the distribution of
a variable of interest in the environment. IPP approaches have been extended to
encompass different sensing modalities (e.g. altitude-dependent sensor models
\cite{popovicetal2020_informativepathplanninga}). Notably, most IPP
approaches consider the spatial phenomenon to be static or at steady-state, or
they assume that the phenomenon does not change meaningfully during the duration
of the survey.

IPP for robotic planning is similar to methods which seek to optimize the
placement or visitation of environmental sensors
\cite{krauseetal2006_nearoptimalsensorplacements}. IPP problems that employ an
adaptive planning approach re-compute vehicle trajectories as observations are
collected. This approach can be framed under the category of problems which
involve sequential decision-making with uncertainty, which in turn can be
formally described as a Partially-observable Markov Decision Process (POMDP)
\cite{kaelblingetal1998_planningactingpartially}. As a constrained optimization
problem, IPP shares may qualities with the orienteering problem
\cite{carpinthayer2022_solvingstochasticorienteering}.

Other methods leverage optimization techniques to determine the most
informative route through a collection of candidate actions or locations.
These approaches include Bayesian optimization
\cite{baietal2016_informationtheoreticexplorationbayesian}, evolutionary
algorithms \cite{popovicetal2017_onlineinformativepath}, and reinforcement
learning \cite{ruckinetal2022_adaptiveinformativepath}.

The asymptotic optimality of rapidly-exploring random trees (RRT) has been
leveraged to solve IPP tasks in a computationally tractable manner, including
exploration applications where the robot is tasked with monitoring an unknown
parameter of interest
\cite{karamanfrazzoli2010_incrementalsamplingbasedalgorithms}.
Rapidly-exploring information gathering (RIG) algorithms approach the IPP task
using incremental sampling with branch and bound optimization
\cite{hollingersukhatme2014_samplingbasedroboticinformation}. Our work builds
on \cite{jadidietal2019_samplingbasedincrementalinformation}, which extended
RIG with an information-theoretic utility function and a related stopping
criterion.

\section{Problem Formulation}
\label{statement}
In this work, we consider the problem of reconstructing a dynamic scalar field
given a limited number of observations, collected along a path. Paths are
generated using a receding-horizon approach, alternating between planning and
execution of the plan until the traveled distance exceeds the budget \(B\) or a
prediction window \(t_{max}\). The task can be formulated as a constrained
optimization problem, where information quantity is to be maximized subject to
an observation cost. In
\cite{hollingersukhatme2014_samplingbasedroboticinformation}, the task is
specified follows:

\begin{equation}
\mathcal{P}^* = \underset{\mathcal{P} \in \Psi}{\textnormal{argmax\ }}
              I(\mathcal{P})\ \textrm{s.t.}\ c(\mathcal{P}) \leq B
\label{ipp-form}
\end{equation}

where \(\mathcal{P}^*\) is an optimal trajectory found in the space of
possible trajectories \(\Psi\), for an individual or set of mobile agents such
that the cost of executing the trajectory \(c(\mathcal{P})\) does not exceed
an assigned motion budget, \(B\). \(I(\mathcal{P})\) is the information gathered
along the trajectory \(\mathcal{P}\), and the movement budget can be any cost
that constrains the effort used to collect observations (e.g., fuel, distance,
time, etc.)

This paper inherits the assumptions of the original RIG formulation and of
prior sampling-based motion planning literature
\cite{hollingersukhatme2014_samplingbasedroboticinformation},
\cite{karamanfrazzoli2010_incrementalsamplingbasedalgorithms} and adds the
following assumptions with respect to time:

\begin{enumerate}
\item The state of the robots and the environment are modeled using discrete time
dynamics
\item Movement of the sampling agent is anisotropic in the time dimension
(see: \autoref{evaluation})
\end{enumerate}

To quantify the information content of a trajectory, we employ a utility
function that optimizes for a reduction in the posterior variance of the GP used
to model the environment. This follows from framing the information gain of an
observation as a reduction of map entropy or uncertainty. In
\cite{bourgaultetal2002_informationbasedadaptive}, the authors present an
approach for quantifying the information content of a map \(M\) as its entropy
\(H\) and the information content of a new observation \(Z\) as the \emph{mutual
information} between \(M\) and \(Z\), denoted as \(I(M;Z)\) and defined as
follows:
\begin{equation}
I(M;Z)=H(M)-H(M \mid Z)
\label{mutual-information}
\end{equation}

We take advantage of the submodularity of mutual information; that is, the information
gained by adding an observation to a smaller set is more useful than adding the
same observation to a larger (super-) set (See
\cite{krauseguestrin2011_submodularityitsapplications} for an analysis of the
benefit of submodular information functions for informative sensing
applications and \cite{guestrinetal2005_nearoptimalsensorplacements} for the
submodularity of mutual information.)

From the perspective of the environmental modeling task, a useful survey is one
that produces the most accurate representation of the environment, minimizing
the expected error given field observations. This follows from equations
\eqref{ipp-form} and \eqref{mutual-information}. This assumption holds when the
model is \emph{well-calibrated} with respect to the priors embodied in the model
parameters \footnote{Refer to Section V and Figure 3 for discussion of the consequences when this assumption does not hold}.
Our approach can be extended to an \emph{adaptive planning} scenario, where model
hyperparameters are updated based on new measurements and future path plans
leverage the updated model. In previous work, we have demonstrated how model
priors can encode modeler intuition, resulting in sampling strategies that vary
in the degree if exploration
\cite{boothcarpin2023_distributedestimationscalar}.
%\vspace{-10pt}

\section{Methods}
\label{methods}
\subsection{Environmental Model}
\label{methods-gpr}
We describe the spatial distribution of an unknown stochastic, dynamic
environmental process occurring in a region \(\xi \subset \mathbb{R}^{2}\) as a
function \(f \colon \mathcal{X} \to \mathbb{R}\) that is sampled and modeled at
the discrete grid, \(\mathcal{X} \subset \mathbb{R}^{N_t \times N_{x,y}}\). Here
\(N_{x,y}\) is a discretization of the spatial domain \(\xi\), while \(N_t\) is
the temporal domain in which the spatial process evolves.

The environmental map comprises this function \(f\) that describes our
observations \(y_i\), plus some additive measurement noise \(\varepsilon_i\),
i.e., \(y_i = f(x_i) + \varepsilon_i\), where we assume that this noise follows an
i.i.d. Gaussian distribution with zero mean and variance \(\sigma_n^2\):
\(\varepsilon \sim \mathcal{N}\left(0, \sigma_{n}^{2}\right)\). We assume that
\(f\) is a realization of a Gaussian process, represented as a probability
distribution over a space of functions. Through marginalization, we can obtain
the conditional density \(f \mid y = \mathcal{N}(\mu_{f \mid y}, \Sigma_{f \mid
y})\). The joint distribution of observations \(\mathbf{y}\), \(\{f(x_1) +
\varepsilon_1, \dots, f(x_n) + \varepsilon_n\}\) and predictions \(\mathbf{f}\),
\(\{f_\star, \dots, f_{\star^n}\}\) at indices \(\mathbf{X_i,t}\),
\(\{x^{(st)}_{1,1}, \dots, x^{(st)}_{m,n}\}\) becomes:

\begin{equation}
  \small
  \begin{bmatrix}
    \mathbf{y} \\
    f(x_{\star}) \\
  \end{bmatrix}
  \sim \mathcal{N} \left(
  0 ,
  \begin{bmatrix}[cc]
  k(\mathbf{X}, \mathbf{X})+\sigma^{2} I_{N} & k\left(\mathbf{X}, x_{\star}\right) \\
  k\left(x_{\star}, \mathbf{X}\right) & k\left(x_{\star}, x_{\star}\right)
  \end{bmatrix} \right)
\label{gp-blockform}
\end{equation}

\noindent
where \emph{s} and \emph{t} denote spatial and temporal indices respectively.
Here, environmental observations \(y\), are drawn from a
training set \(\mathcal{D}\) of \(n\) observations,
\(\mathcal{D}= (X,\mathbf{y}) = \left\{\left(\mathbf{x}_{i,t}, y_{i,t}\right) \mid
i=1, \ldots, n\right\}\). \(k\) is the covariance function (or kernel),
\(\sigma^2_n\) is the variance of the observation noise, and input vectors
\(\mathbf{x}\) and query points \(\mathbf{x_\star}\) of dimension \(D\),
are aggregated in the \(D \times n\) design matrices \(X\) and \(X_\star\)
respectively. From the Gaussian process, we can obtain estimations of both
the expected value of the environmental field and the variance of each
prediction. Noteworthy is the posterior variance, which takes the form:
\begin{align}
\label{posterior-variance}
\sigma = \mathbb{V}\left[f_\star\right] &=
  k \left(x_\star, x_\star\right) -
  k \left(x_\star, \mathbf{X}\right) \times \\
  &\quad \left[k(\mathbf{X}, \mathbf{X}) +
    \sigma_{n}^{2} \mathbf{I}_{n}\right]^{-1}
    k\left(\mathbf{X}, \mathbf{x}_{*}\right) \nonumber
\end{align}

The differential entropy of a Gaussian random variable is a monotonic function of
its variance, and can be used to derive the information content of a proposed
measurement. We will show how this can be used to approximate information gain
(equation \eqref{mutual-information}) in \autoref{methods-infofun}.

It is important to note that for fixed kernels the variance does not depend on
the value of the observation, allowing us to reason about the effectiveness of a
proposed observation before traveling to the sampling location
\cite{krauseetal2008_nearoptimalsensorplacements}. Also notable is the kernel
\emph{k} which establishes a prior over the covariance of any pair of observations.
Separate priors can be established in spatial or temporal dimensions, leading to
the opportunity to incorporate spatial and/or temporal domain knowledge into the
planning process.

\subsection{Spatiotemporal prior}
\label{methods-prior}
The modeling effort can be framed as a multi-task (or multi-output) prediction
of correlated temporal processes at each spatial discretization \(N_{x,y}\). As
we only have a finite set of sampling vehicles (one, in fact), we cannot observe
all of the spatial "outputs" for a given time, however we can establish a basis
upon which they can be correlated
\cite{goovaertsgoovaerts1997_geostatisticsnaturalresources}. Specifically, the
Linear Model of Coregionalization (LMC) has been applied to GP regression where
\emph{p} outputs are expressed as linear combinations of independent random
vector-valued functions \(f: \mathcal{T} \rightarrow \mathbb{R}^{p}\). If
these input functions are GPs, it follows that the resulting model will also be
a GP \cite{alvarezlawrence2011_computationallyefficientconvolved}. The
multi-output GP (MOGP) can be described by a vector-valued mean function and a
matrix-valued covariance function (see Equation \eqref{posterior-variance}). A
practical limitation of MOGPs has been their computational complexity. For
making \(p\) predictions with \(n\) input observations \(y\left(t_{1}\right), \ldots, y \left(t_{n}\right) \in \mathbb{R}^{p}\), the
complexity of inference is \(\mathcal{O}\left(n^{3} p^{3}\right)\) in time and
\(\mathcal{O}\left(n^{2} p^{2}\right)\) in memory
\cite{bruinsmaetal2020_scalableexactinference}. A variety of strategies exist
to solve lighter, equivalent inference tasks under simplifying assumptions, such
as expressing an output from linear combinations of latent functions that share
the same covariance function, but are sampled independently
\cite{alvarezlawrence2011_computationallyefficientconvolved}. Since our
information function is only dependent on the posterior covariance, we can take
advantage fast approximations with complexity \(\mathcal{O}(k(n+p \log p)\)
(see discussion in \autoref{methods-infofun}).

As mentioned earlier, the kernel \(k\) establishes a prior likelihood over the
space of functions that can fit observed data in the regression task. For the
regression of discretely-indexed spatiotemporal data, where space is indexed by
\(s\) (eg. latitude/longitude) and time is indexed by \(t\) (eg. seconds), we build a
composite kernel by multiplying a spatial and temporal kernel:

\vspace{-10pt}
\begin{equation}
k((s, t),t(s^{\prime}, t^{\prime}))=k_{s}(s, s^{\prime}) k_{t}(t, t^{\prime})
\label{composite-kernel}
\end{equation}

While other approaches to kernel composition are possible and encode different
environmental priors, constructing a kernel that is separable along input
dimensions affords considerable computational advantages. More generally,
when \(k(\mathbf{x}, \mathbf{x}^{\prime}) =
\prod_{d=1}^{D} k^{(d)}(\mathbf{x}^{(d)}, \mathbf{x}^{\prime(d)})\), the kernel
(Gram) matrix \(K\) can be decomposed into smaller matrices \(K=K_1 \otimes
\cdots \otimes K_D\) which can be computed in
\(\mathcal{O}(D n^{\frac{D+1}{D}})\) time (see
\cite{wilsonetal2014_fastkernellearning} and
\cite{flaxman2015_machinelearningspace} for more on kernel composition for
multidimensional regression.)

For the spatial relation, we use the Matérn kernel with \(\nu = 3/2\) and fixed
hyperparameters. Comprehensively described in
\cite{stein1999_interpolationspatialdata}, the Matérn kernel is a
finitely-differentiable function with broad use in the geostatistical literature
for modeling physical processes due in part to its ability to resist
over-smoothing natural phenomena with sharp discontinuities. It takes the form:

\vspace{-10pt}
\begin{equation}
K_{\text {Matern}}(X, X_\star) =
  \sigma^2 \frac{2^{1- \nu}}{\Gamma(\nu)}
  \left(\frac{\sqrt{2 v}}{l} r\right)^{\nu}
  K_{\nu}\left(\frac{\sqrt{2 \nu}}{l} r\right)
\label{cov-matern}
\end{equation}

\noindent \\[0pt]
where \(K_{\nu}\) is a modified Bessel function , \(\Gamma(\cdot)\) is the
Gamma function, and \(r\) is the Euclidean distance between input points \(X\)
and \(X_\star\). \(\nu > 0\), \(l > 0\), and \(\sigma^2 > 0\) are
hyperparemeters representing smoothness, lengthscale, and observation variance
respectively. We use a radial basis function kernel (RBF or squared-exponential)
in the time dimension to smoothly capture diffusive properties that may fade in
time. Note that the Matérn kernel approaches the RBF as \(\nu \to \infty\).

\subsection{Informative Planning}
\label{methods-ipp}
In this work, we present a novel planner \(\texttt{IIG-ST}\) to address IPP task
defined in equation \eqref{ipp-form}. Our planner is built upon
\(\texttt{IIG-Tree}\), a sampling-based planner with an information-theoretic
utility function and convergence criterion
\cite{jadidietal2019_samplingbasedincrementalinformation} and derived from the
family of Rapidly-exploring Information Gathering (RIG) algorithms introduced by
Hollinger and Sukhatme
\cite{hollingersukhatme2014_samplingbasedroboticinformation}. RIG inherits the
asymptotic cost-optimality of the \(\mathrm{RRT}^\star\), \(\mathrm{RRG}\), and
\(\mathrm{PRM}^\star\) algorithms
\cite{karamanfrazzoli2011_samplingbasedalgorithmsoptimal}, a conservative
pruning strategy from the branch and bound technique
\cite{binneysukhatme2012_branchboundinformative}, and an information-theoretic
convergence criterion (see discussion in \autoref{methods-convergence}). We add
routines to consider the time dimension of samples in the tree and combine it
with a hybrid covariance function and stopping criterion grounded in map
accuracy.

\subsection{Information Functions}
\label{methods-infofun}
From equation \eqref{mutual-information}, we established information gain as the
reduction of map entropy \(H\) given a new observation \(Z\).

If the map is modeled as a Gaussian Process where each map point (or query
point) is a Gaussian random variable, we can approximate mutual entropy with
differential entropy. For a Gaussian random vector of dimension \(n\),
the differential entropy can be derived as
\(h(X)=\frac{1}{2} \log \left((2 \pi e)^{n}|\Sigma|\right)\).
If we let \(X \sim \mathcal{N}\left(\mu_{X}, \Sigma_{X}\right)\) and
\(X \mid Z \sim \mathcal{N}\left(\mu_{X \mid Z}, \Sigma_{X \mid Z}\right)\)
be the prior and posterior distribution of the random vector \(X\), before
and after incorporating observation \(Z\), then the mutual information
becomes:
\begin{equation}
I(X;Z)=\frac{1}{2} \left[\log \left(|\Sigma_{X}|\right) -
                         \log \left(|\Sigma_{X \mid Z}|\right) \right]
\label{mutual-information-logcov}
\end{equation}
where \(\Sigma\) is the full covariance matrix.

For a random vector \(\mathbf{X} = (X_1,\ldots,X_n)\) with covariance matrix
\(\textbf{K}\), the mutual information between \(\mathbf{X}\) and observations Z
can be approximated from equation \eqref{mutual-information-logcov} as:

\begin{equation}
\hat{I}(X;Z)=\sum_{i=1}^{n} \frac{1}{2} \left[\log \left(\sigma_{X_{i}}\right) -
                                             \log \left(\sigma_{X_{i} \mid Z}\right)\right]
\label{mutual-information-approx}
\end{equation}

Using marginalization, for every
\(X_i\), it holds that \(\mathbb{V}\left[X_{i}\right]= K^{[i, i]}\). The expression becomes:

\begin{equation}
\hat{I}^{[i]}\left(X_{i} ; Z\right)=\frac{1}{2}\left[\log \left(\sigma_{X_{i}}\right)-\log \left(\sigma_{X_{i} \mid Z}\right)\right]
\end{equation}

and can be computed as the sum of marginal variances at \emph{i}:
\(\hat{I}(X ; Z)=\sum_{i=1}^{n} \hat{I}^{[i]}(X_{i} ; Z)\)
(see \cite{jadidietal2019_samplingbasedincrementalinformation} for a
derivation).

The main motivation of using marginal variances at evaluation points (Equation
\eqref{mutual-information-approx}) is to avoid maintaining and updating
(inverting) the full covariance matrix. This is of a particular concern for
spatiotemporal modeling, because the number of inducing points grows on the
order of \(m \times n\) for a spatial domain of \(m\) rows and \(n\) columns.
Alternate GP formulations such as spatio-temporational sparse variational GPs
(ST-SVGP) allow for computational scaling that is linear in the number of time
steps \cite{hamelijncketal2021_spatiotemporalvariationalgaussiana} For
computing the posterior variance at GP inducing points, we use LOVE (LanczOs
Variance Estimates), for a fast, constant-time approximation of predictive
variance
\cite{pleissetal2018_constanttimepredictivedistributions,gardneretal2018_gpytorchblackboxmatrixmatrix}.

\begin{algorithm}[ht]
\caption{Information\_GPVR-ST()}
\begin{algorithmic}[1]
  \Require
    \Statex Proposed robot pose or location from RRT/RIG \texttt{Steer} \( p \),
    current map/state estimate \( \mathcal{M}_\mathcal{D} \), covariance function \( k(\cdot, \cdot) \),
    prior map variance \( \sigma \), variance of observation noise \( \sigma_{n}^{2} \),
    near node information \( I_{\text{near}} \);
  \State\( \bar{\sigma} \leftarrow \sigma \)
  \Comment{Initialize updated map variance as the current map variance}
  \If{\( I_{\text {near }} \) is not empty}
    \Comment{Initialize information gain}
    \State \( I \leftarrow I_{\text {near }} \)
  \Else
    \State \( I \leftarrow 0 \)
  \EndIf
  \State \( z \leftarrow \) Propose a future measurement at location \( p \) and map \( \mathcal{M} \)
  \Comment{Calculate posterior map variance at training and query points}
  \State \( \bar{\sigma}  \leftarrow  \texttt{LOVE}\left(X, X_{*}\right) \)
  \ForAll{\( i \in \mathcal{M}_\mathcal{D} \)}
    \State \( I \leftarrow I + 1/2\left[ \operatorname{logdet}\left(\sigma^{[i]}\right) -
                                         \operatorname{logdet}\left(\bar{\sigma}^{[i]}\right)\right] \)
  \EndFor
  \State \Return \( I \) (total information gain), \( \bar{\sigma} \) (updated map variance)
\end{algorithmic}
\label{info-gpvr}
\end{algorithm}
Algorithm \ref{info-gpvr} details the procedure for updating a node's information
content. In lines 6-8, the location of a future measurement \(z\) at pose \(p\),
is added to the set of past observations (training points) from the entire node
graph. This is used to create a new map state containing the previous training
points plus the new measurement and the preexisting query points where the GP is
evaluated. Next, the posterior variance is calculated (lines 8) using LOVE
(LanczOs Variance Estimates)
\cite{pleissetal2018_constanttimepredictivedistributions,gardneretal2018_gpytorchblackboxmatrixmatrix}
to produce a posterior variance at the proposed locations of training points \(X
\in \mathcal{M}_\mathcal{D}\), query points \(X_{*} \in
\mathcal{M}_\mathcal{D}\), and the variance of observation noise
\(\sigma_{n}^2\). Finally, information content of the entire posterior map is
updated and the information gain is returned as a marginal variance (lines
9-11).

\subsection{Convergence criterion}
\label{methods-convergence}
The closely related Incrementally-exploring Information Gathering (IIG)
algorithm modifies RIG with an information-theoretic convergence criterion
\cite{jadidietal2019_samplingbasedincrementalinformation}. Specifically, IIG
bases the stopping criterion around a \emph{relative information contribution} (RIC)
criterion that describes the marginal information gain of adding a new
observation relative to the previous state the RIG tree (see Equation 15 in
\cite{jadidietal2019_samplingbasedincrementalinformation} for a comprehensive
discussion of the IIG algorithm and for a definition of the RIC). There, it was
used as a tunable parameter that established a planning horizon for information
gathering. In this paper, we use posterior map variance as a lower bound for
mean-square error (MSE) (Equation \eqref{mse}) at a arbitrary test location in the
GP, given optimal hyperparameters \(\theta\) for the GP regression model. We
replace the stopping criterion in IIG with a threshold established by the
operator as the lower bound of expected prediction MSE.
\begin{equation}
\operatorname{MSE}\left(\widehat{f_{\star}}\right) \geq
   \underbrace{
      \mathbb{V}\left[f_\star\right]
              }_{=\sigma_{\star \mid y}^{2}(\theta)}
\label{mse}
\end{equation}

It is important to note that this inequality holds for the hyperparameters
\(\theta\) that produce an optimal predictor of \(f\) (see Result 1 in
\cite{wagbergetal2017_predictionperformancelearning} for a proof of Equation
\eqref{mse} using the Bayesian Cramér-Rao Bound (BCRB).) In practice, \(\theta\) is
learned from the data. For approximate (suboptimal) values of \(\theta\), the
bound of Equation \eqref{mse} will not hold, as additional error is introduced
from the unknown model hyperparameters. However, when coupled with adaptive
planning techniques to learn \(\theta\) from observations, then the posterior
variance approaches the true lower bound of the MSE. A deeper analysis of the
implications of this application is a target of future work.

\subsection{Path selection and planning}
\label{methods-pathselect}
Once the planner terminates (either by the convergence criterion or after a
fixed planning horizon), a path must be selected from the graph of possible
sampling locations. We use a vote-based heuristic from
\cite{jadidietal2019_samplingbasedincrementalinformation} that ranks paths
according to a similarity ratio and biases towards paths that are longer and
more informative with a \emph{depth-first search}. In the simulated environment,
parameters are set for vehicle speed, sampling frequency, and replanning
interval. The vehicle alternates between planning, executing, and replanning
in a receeding-horizon fashion, such that 2-3 waypoints are visited in each
planning interval.

The path selection strategy is independent of the informative path planning
algorithm and can be thought of as an orienteering problem within a tree of
sampling locations.

% Figure caption space is 10pt by default, here we narrow it to 5
\begin{figure}[h]
    \centering
    \includegraphics[width=\columnwidth]{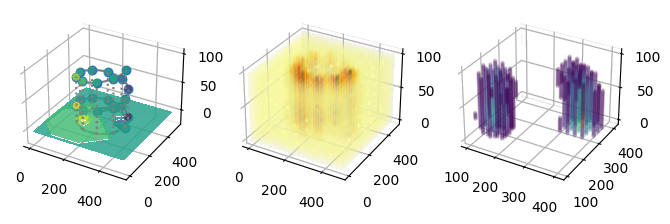}
    \includegraphics[width=\columnwidth]{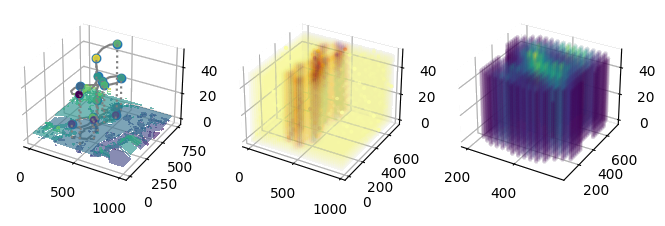}
    \vspace{-20pt}
    \caption{A visualization of the benchmark (coverage) sampling scenarios
    (top: fluid simulation, bottom: ocean sampling simulation).
    The posterior variance is depicted in
    the second panel, and the posterior mean in the third, with near-zero values
    filtered show the underlying structure. The coverage planners are given a
    path budget and node budget equivalent to the median of the equivalent metrics
    among all runs of the informed planners. Observations are collected on a
    circlular coverage in the synthetic environment and a lemniscatic coverage in
    the oceanic experiment.
    }
    \label{fig2}
\vspace{-10pt}
\end{figure}

\section{Experimental Evaluation and Discussion}
\label{evaluation}
In this section, we contrast our proposed spatiotemporal-informed planner
(IIG-ST) against a traditional coverage survey strategy (see Figure \ref{fig2}),
and an informed planner that does not consider temporal variation (IIG). We
evaluate the accuracy of the final map representation at the end of the survey
period under varying choices of spatial and temporal priors. We also consider
the ancillary objective of making predictions of the state of environment at
arbitrary points in time. This can be useful for objectives that wish to
reconstruct the dynamics of a system, such as modeling a vector field. However,
this is complicated by the fact that the survey envelope is anisotropic in the
temporal dimension -- the robot and sensor can only travel forward through time.

% Figure caption space is 10pt by default, here we narrow it to 5
\begin{figure*}
    \centering
    \includegraphics[width=\textwidth]{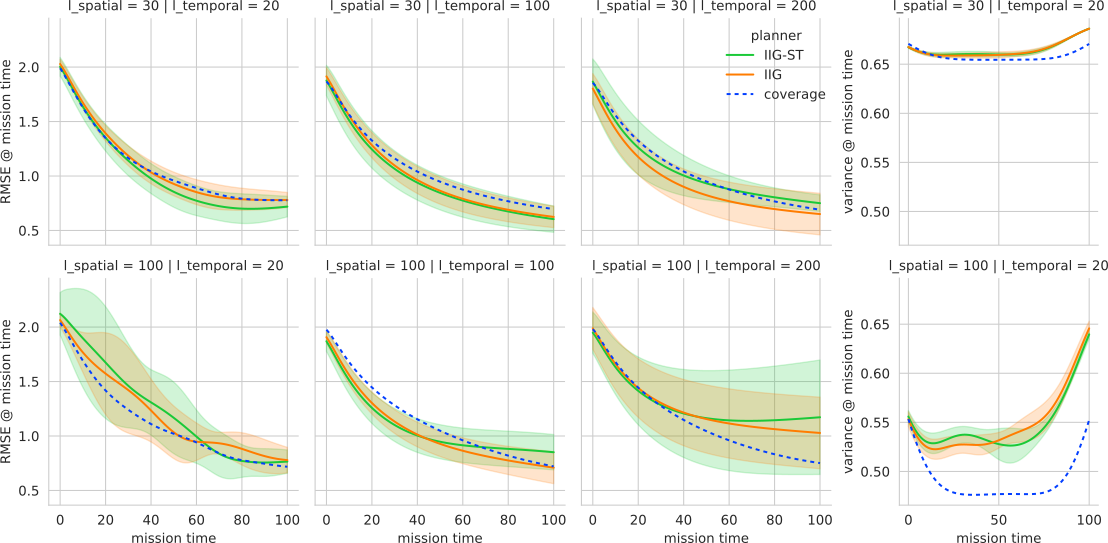}
    \vspace{-10pt}
    \caption{[Advection/diffusion simulation] A comparison of map error and
    posterior variance (lower is better) at different locations in the mission time for different spatiotemporal priors
    Optimal priors are chosen in the top left panel (\(\ell_t = 20\) and \(\ell_s = 30\))
    and become increasingly suboptimal in other panels.
    IIG-ST (our planner) is compared the same planner lacking time information (IIG)
    and a circular survey strategy. The error metric is expressed across the entire
    spatial domain at different time indices (denoted on the x-axis), and reflects
    the error between the estimated map and the state of the environment
    \emph{at that time}. Y-axis scales are shared between rows.
    }
    \label{fig3}
\vspace{-10pt}
\end{figure*}

\subsection{Experimental setting}
\label{sec:org8b7238c}
Our objective is to model the end-state of a spatial phenomenon that undergoes
advection and diffusion in a 2D environment. This can represent the movement of
a substance of interest in a fluid, a porous medium such as soil, or any number
of similar natural processes. Two fluid parcels are initialized with
inversely-proportional velocities, at opposite corners of a \(500 \times
500\)-unit gridded environment. The fluid parcels advect and diffuse according
to the Navier-Stokes equations for an incompressible fluid, implemented as a
forward-differencing discretization without boundary conditions.

We initialized the RIG-planner with fixed planning parameters: the vehicle can
move a maximum of 100 map-units, every 5 time-units. Replanning is done every 10
time increments, and planning within each increment stops when estimated
\(\mathbb{V}\left[f_\star\right] = 0.15\). Sampling occurs once every 5 time
increments. We set the time budget to be 100 units and compute the accuracy of
the final representation of the map at \(t = 50\) min. Map accuracy at different
moments in mission time are presented in Figure \ref{fig3}. While the planner was
not given a movement budget, the fixed speed of the vehicle and finite
time-horizon resulted in consistent numbers of observations \((M = 21.0, SD =
0.2)\) and path lengths \((M = 1236, SD = 36)\) among the informative planners.
The coverage baseline is given a proportional budget (21 observations, 1610 map
units traveled). This is sufficient to complete a full tour of the environment
with revisitation (see Figure \ref{fig2}). The full table of parameters set for
the planner can be found in the accompanying video. We executed the experiments
in a GNU/Linux environment on a 3.6 GHz Intel i7-4790 computer with 11 GB of RAM
available. All procedures used single-threaded Python implementations for RRT
sampling from \cite{sakaietal2018_pythonroboticspythoncode} and multi-threaded
posterior variance final map predictions were performed using implementations
from GPyTorch \cite{gardneretal2018_gpytorchblackboxmatrixmatrix} without GPU
or TPU acceleration so as to simulate the resources available on an embedded
system.

\subsection{Consequences of the temporal prior}
\label{sec:org26f039b}

To demonstrate the consequences of incorporating a spatiotemporal prior on
informative planning in dynamic fields, we use the composite
covariance function given in equation \eqref{composite-kernel} both in planning
and for evaluating the accuracy of the final map representation. This is
notable for the baseline comparisons--while the coverage planner follows a
deterministic trajectory, different map accuracies and variance reductions are
expected depending on the choice of spatiotemporal prior during the construction
of the final map model.

For the temporal relation, we use a RBF kernel with length scales of \(\ell_t =
20, 100, 200\) time units. The spatial relation comprises a Matérn kernel with
\(\nu = 3/2\) and length scales of \(\ell_s = 100\) distance units. To verify
that the robot solves the problem in \autoref{statement}, we evaluate the
root-mean squared error between the map representation at \(t = 100\) and the
state of the field at the same time. As the planner only requires the posterior
covariance, it is not necessary to produce continuous estimations of the map
state, so the final representation is computed once the simulation has ended. 20
episodes are run for each hyperparameter combination and summaries of average
error, average posterior variance and standard deviations are found in table
\ref{tab1}.

\begin{table*}
\begin{minipage}{.65\textwidth}
\scriptsize
\centering
\begin{tabular}{l|ll|cc|cc}
 &  &  & \multicolumn{2}{c}{\emph{RMSE}} &  \multicolumn{2}{|c}{\(\overline{V}\)} \\
\midrule
planner & \(\ell_s\) & \(\ell_t\)  & \(t_{max}\) & \(t_{all}\) &  \(t_{max}\) &   \(t_{all}\)  \\
\midrule
\multirow[c]{4}{*}{IIG} & \multirow[c]{2}{*}{30} & 20 & 0.781 (\emph{0.066}) & 1.123 (\emph{0.072}) & 0.686 (\emph{0.0}) & 0.664 (\emph{0.001}) \\
 &  & 100 & 0.612 (\emph{0.096}) & 1.035 (\emph{0.098}) & 0.64 (\emph{0.005}) & 0.626 (\emph{0.006}) \\
 & \multirow[c]{2}{*}{100} & 20 & 0.762 (\emph{0.113}) & 1.288 (\emph{0.179}) & 0.645 (\emph{0.007}) & 0.547 (\emph{0.004}) \\
 &  & 100 & 0.75 (\emph{0.222}) & 1.093 (\emph{0.116}) & 0.462 (\emph{0.02}) & 0.413 (\emph{0.025}) \\
 \midrule
\multirow[c]{4}{*}{IIG-ST} & \multirow[c]{2}{*}{30} & 20 & 0.733 (\emph{0.089}) & 1.092 (\emph{0.064}) & 0.686 (\emph{0.0}) & 0.665 (\emph{0.001}) \\
 &  & 100 & 0.611 (\emph{0.121}) & 1.028 (\emph{0.135}) & 0.638 (\emph{0.005}) & 0.624 (\emph{0.006}) \\
 & \multirow[c]{2}{*}{100} & 20 & 0.768 (\emph{0.101}) & 1.3 (\emph{0.238}) & 0.64 (\emph{0.004}) & 0.547 (\emph{0.005}) \\
 &  & 100 & 0.866 (\emph{0.194}) & 1.114 (\emph{0.117}) & 0.458 (\emph{0.014}) & 0.414 (\emph{0.017}) \\
 \midrule
\multirow[c]{4}{*}{coverage} & \multirow[c]{2}{*}{30} & 20 & 0.777 & 1.132 & 0.671 & 0.658 \\
 &  & 100 & 0.697 & 1.099 & 0.639 & 0.638 \\
 & \multirow[c]{2}{*}{100} & 20 & 0.718 & 1.173 & 0.552 & 0.491 \\
 &  & 100 & 0.721 & 1.19 & 0.398 & 0.394 \\
\bottomrule
\end{tabular}
\end{minipage}%
\begin{minipage}{.2\textwidth}
\scriptsize
\centering
\renewcommand{\arraystretch}{1.3}
\begin{tabular}{l|l|cc}
 &   &  \multicolumn{2}{|c}{\emph{RMSE}} \\
\midrule
planner & \(\ell_s\) & \(t_{max}\) & \(t_{all}\) \\
\midrule
\multirow[c]{3}{*}{IIG} & 5 & 6.654 (\emph{0.015}) & 5.499 (\emph{0.004}) \\
 & 40 & 5.934 (\emph{0.382}) & 4.926 (\emph{0.087}) \\
 & 100 & 3.835 (\emph{0.725}) & 3.777 (\emph{0.17}) \\
\midrule
\multirow[c]{3}{*}{IIG-ST} & 5 & 6.658 (\emph{0.013}) & 5.499 (\emph{0.003}) \\
 & 40 & 5.846 (\emph{0.305}) & 4.904 (\emph{0.072}) \\
 & 100 & 4.1 (\emph{0.739}) & 3.698 (\emph{0.252}) \\
\midrule
\multirow[c]{3}{*}{coverage} & 5 & 3.826 & 6.238 \\
 & 40 & 3.281 & 5.506 \\
 & 100 & 2.909 & 4.56 \\
\bottomrule
\end{tabular}
\end{minipage}
\caption{(L) [Advection/diffusion] Aggregated (\(n=20\)) map accuracy (RMSE)
and posterior variance (mean, \emph{std}) of the spatiotemporal planner (IIG-ST)
compared to a spatial-only and deterministic survey strategies for fixed length scales.
(R) [Ocean dataset] Aggregated \(n=20\) map accuracy for the ocean water quality
experiment (\(\ell_t = 100\) for all runs). Lower numbers are better. Note: standard
deviation values are not expressed for the deterministic planner.}
\label{tab1}
\vspace{-15pt}
\end{table*}

In Figure \ref{fig3}, we examine the choice of kernel hyperparameters on the
performance of our planner. Optimal parameters were established offline using
the baseline samples and a standard marginal log likelihood function and the
Adam optimizer in gpytorch (\(\ell_t = 20\) and \(\ell_s = 30\)). These serve as
the basis of comparison in the top-left panel of Figure \ref{fig3} and resulted
the spatiotemporal planner outperforming the temporally-naive and baseline
planner for on average, throughout the entire mission duration. Large
lengthscales imply a greater degree of correlation across space or time, and
result a greater reduction of posterior variance. A reduction of model
uncertainty should translate to a higher map accuracy, however this is not the
case if the spatial priors are unrepresentative. For example, while the coverage
planner had lower variance due to a longer path traveled and more dispersed
observations, the resulting map accuracy was not better than the informative
planners, leading to the conclusion that the spatiotemporal prior did not
reflect the variation of the observed process. We want to emphasize that path
planning algorithms based around variance reduction should also place the metric
within a broader context of the practical objective -- map accuracy.

% Figure caption space is 10pt by default, here we narrow it to 5
\begin{figure}[h]
    \centering
    \includegraphics[width=\columnwidth]{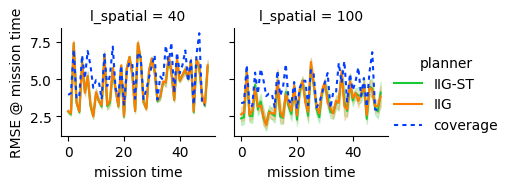}
    \includegraphics[width=0.5\columnwidth]{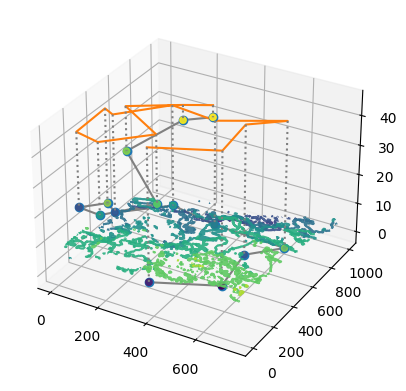}
    \includegraphics[width=0.4\columnwidth]{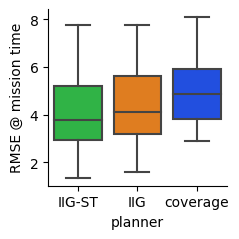}
    \vspace{-5pt}
    \caption{Example results from the ocean modeling experiments.
    (Top) Map error as a function of mission time, (\(\ell_t = 100\)).
    (L) Example trajectory, with path trace projected above a
    representation of the environment at \(t=0\). (R) Aggregated statistics
    from the figures in the top panel.}
    \label{fig4}
    \vspace{-10pt}
\end{figure}

For informative planners, the effect is magnified, as the planner will move
toward more dispersive sampling, thus missing high-frequency spatial phenomena
entirely. This is demonstrated in the marginally improved accuracy and lower
posterior variance for IIG-ST when given a unrepresentative spatial and temporal
prior. In worst-case scenarios, a very unrepresentative temporal prior (\(\ell_t
= 200\)) can reduce the performance of the spatiotemporal planner \emph{below} the
baseline (Figure \ref{fig3}, Col. 2). As the ultimate goal of informed robotic
sensing \emph{is} model accuracy and not simply variance reduction, hyperparameter
optimization must be a key component for accurate mapping and is a common
practice in adaptive planning
\cite{fuhgetal2021_stateoftheartcomparativereview}. Furthermore, a time-varying
kernel could be specified and optimized as observations of the environment are
gathered. Future work will investigate the effect and performance of updating
model priors during the course of a survey mission.

The final map posterior is evaluated with the same spatiotemporal kernel in all
cases, regardless of planning method to ensure a fair comparison between the
methods. Only the spatiotemporal planner (IIG-ST) is able to make use of
temporal variance during replanning. Training observations are obtained from a
point sensor model, where the a "sample" is obtained by the simulated agent
querying the ground-truth scalar field at a sample location. We use a sparse
representation of posterior variance, evaluated at a 1/20 scale spatial
resolution for a total of \(25 \times 25 \times 50\) query (inducing) points.
Recent advancements in spatiotemporal GPs with separable kernels, enable
computational scaling to scale lineally in the temporal dimension, instead of
cubic \cite{hamelijncketal2021_spatiotemporalvariationalgaussiana}. These and
other recent developments are reducing the computational burden of large GPs and
informative planning with spatiotemporal information at a large scale.

\subsection{Ocean particulate mapping scenario}
\label{sec:org11a2fe9}

We demonstrate our spatiotemporal IPP approach in a syoptic-scale simulation
using real-world ocean reflectance data. The data was collected in an
approximately 1500 x 1000 \emph{km} region off the west coast of California from the
Moderate Resolution Imaging Spectroradiometer (MODIS) aboard NASA's Terra and
Aqua earth observation satellites
\cite{nasaoceanbiologyprocessinggroup2017_modisaqualevelmapped}. Rasters of
weekly median reflectance from band 9 (443 \emph{nm} wavelength) were assembled for
the calendar year of 2020. Backscattered light in this wavelength band is highly
correlated with the concentration of suspended organic and inorganic particles
(e.g. sediments) in the water. In terrestrial and oceanic waters, this can be
used as an indicator of water quality, which can guide management decisions
related to water diversion and treatment.

We simulated an autonomous aquatic vehicle (AUV) with characteristics similar to
the Wave Glider, which is an AUV capable of extended oceanic monitoring
campaigns by using oceanic waves for propulsion. Based on the long-mission
average speed of 1.5 knots, our simulated vehicle could cover a maximum of 330
km per week. We compare the performance of our informed planner against a fixed
lemniscatic coverage pattern. As with the previous section, we evaluate the RMSE
of the map representation, both at the final time step and at arbitrary temporal
increments in the mission envelope. Summaries of average error, standard
deviations, and posterior variance are presented in Table \ref{tab1} and Figure
\ref{fig4}. As with the previous experiment, posterior variance and map accuracy
are evaluated at a 1/20 scale spatial resolution. Also, as with the previous
experiment, the performance of IIG-ST is sensitive to the choice of
hyperparameters.

\section{Conclusion}
\label{conclusion}
This work presented an approach for environmental modeling using a novel
spatiotemporally-informed path planner. We presented a framework for quantifying
the information gain of sampling locations based on their location and time and
quantifying the operative outcome -- map accuracy. We show that this informed
strategy is computationally tractable with modern computational techniques and
can outperform naive and conventional approaches, conditional on an appropriate
spatiotemporal prior.
Multiple avenues for future work lead from this effort. Adaptive planning can be
used to revise the spatiotemporal prior as measurements are collected between
replanning intervals. This approach can be extended to consider time-varying
kernels. variable sensor models, and multi-robot systems.

\bibliographystyle{plain}
\bibliography{00_draft_ipp}
\end{document}